\documentclass[letterpaper, 10pt, conference]{ieeeconf}  

\IEEEoverridecommandlockouts                              

\usepackage{dblfloatfix}
\usepackage{graphicx} 
\usepackage{amsmath} 
\usepackage{amssymb}  
\usepackage{kantlipsum}
\usepackage{balance}
\usepackage{multirow}
\usepackage{cite}
\usepackage[normalem]{ulem} 
\usepackage{mathtools}
\usepackage[ruled,linesnumbered]{algorithm2e}
\usepackage{tabulary,booktabs}
\usepackage{subfigure}
\usepackage{algpseudocode}

\title{\Large \bf Cross Domain Robot Imitation with Invariant Representation}
\author{Zhao-Heng Yin$^{1}$, Lingfeng Sun$^{3}$, Hengbo Ma$^{3}$, Masayoshi Tomizuka$^3$, Wu-Jun Li$^{2}$ 
\thanks{$^{1}$Zhao-Heng Yin is with the Department of Electronic and Computer Engineering, The Hong Kong University of Science and Technology, Hong Kong SAR. \texttt{zhaohengyin@gmail.com}}%
\thanks{$^{2}$Wu-Jun Li is with the Department of Computer Science and Technology, Nanjing University, Nanjing 210012, PRC. \texttt{liwujun@nju.edu.cn}}%
\thanks{$^{3}$Lingfeng Sun, Hengbo Ma and Masayoshi Tomizuka are with the Department of Mechanical Engineering, University of California, Berkeley, Berkeley, CA 94720, USA. 	\texttt{\{lingfengsun, hengboma, tomizuka\}@berkeley.edu}}%
}

\begin{document}
\maketitle
\begin{abstract}
    Animals are able to imitate each others' behavior, despite their difference in biomechanics. In contrast, imitating the other similar robots is a much more challenging task in robotics. This problem is called cross domain imitation learning~(CDIL). In this paper, we consider CDIL on a class of similar robots. We tackle this problem by introducing an imitation learning algorithm based on invariant representation. We propose to learn invariant state and action representations, which aligns the behavior of multiple robots so that CDIL becomes possible. Compared with previous invariant representation learning methods for similar purpose, our method does not require human-labeled pairwise data for training. Instead, we use cycle-consistency and domain confusion to align the representation and increase its robustness. We test the algorithm on multiple robots in simulator and show that unseen new robot instances can be trained with existing expert demonstrations successfully. Qualitative results also demonstrate that the proposed method is able to learn similar representations for different robots with similar behaviors, which is essential for successful CDIL.
\end{abstract}

\section{Introduction}
Animals are able to imitate each others' behavior by watching their demonstrations despite their difference in biomechanics such as body length, shape, and strength~\cite{zentall2001imitation}. However, previous robotics research suggests that imitating a similar reference robot of different embodiment and dynamics is a challenging problem, which is also termed as Cross Domain~(robot) Imitation Learning~(CDIL) in the literature~\cite{adapt_dail}. Solving this problem can be both appealing and significant for robotics. On one hand, domain discrepancies between the expert and the agent usually exist in the practice of imitation learning, and direct imitation by Behavior Cloning~(BC)~\cite{Behavior_cloning} or Inverse Reinforcement Learning~(IRL)~\cite{IRL} without noticing such discrepancies will harm learning performance~\cite{adapt_dail}. On the other hand, CDIL will also make imitation learning more flexible and convenient, as this enables us to use existing massive amount of demonstrations online to train similar robots. 

Previous CDIL research usually focuses on cross domain imitation learning between a pair of robots~\cite{adail, tpil}. But the demonstrations can actually come from multiple sources (experts) in practice. Therefore in this paper, we consider a more general CDIL problem among a class of similar robots. One example is shown in the Figure 1, which is taken from the MuJoCo Walker problem~\cite{mujoco}. We are provided with demonstration of several walkers of different embodiments~(leg length), and our goal is to use these demonstrations to train a new Walker instance sampled from a class of Walker robots. 

\begin{figure}[t]
\centering
\includegraphics[width=0.85\linewidth]{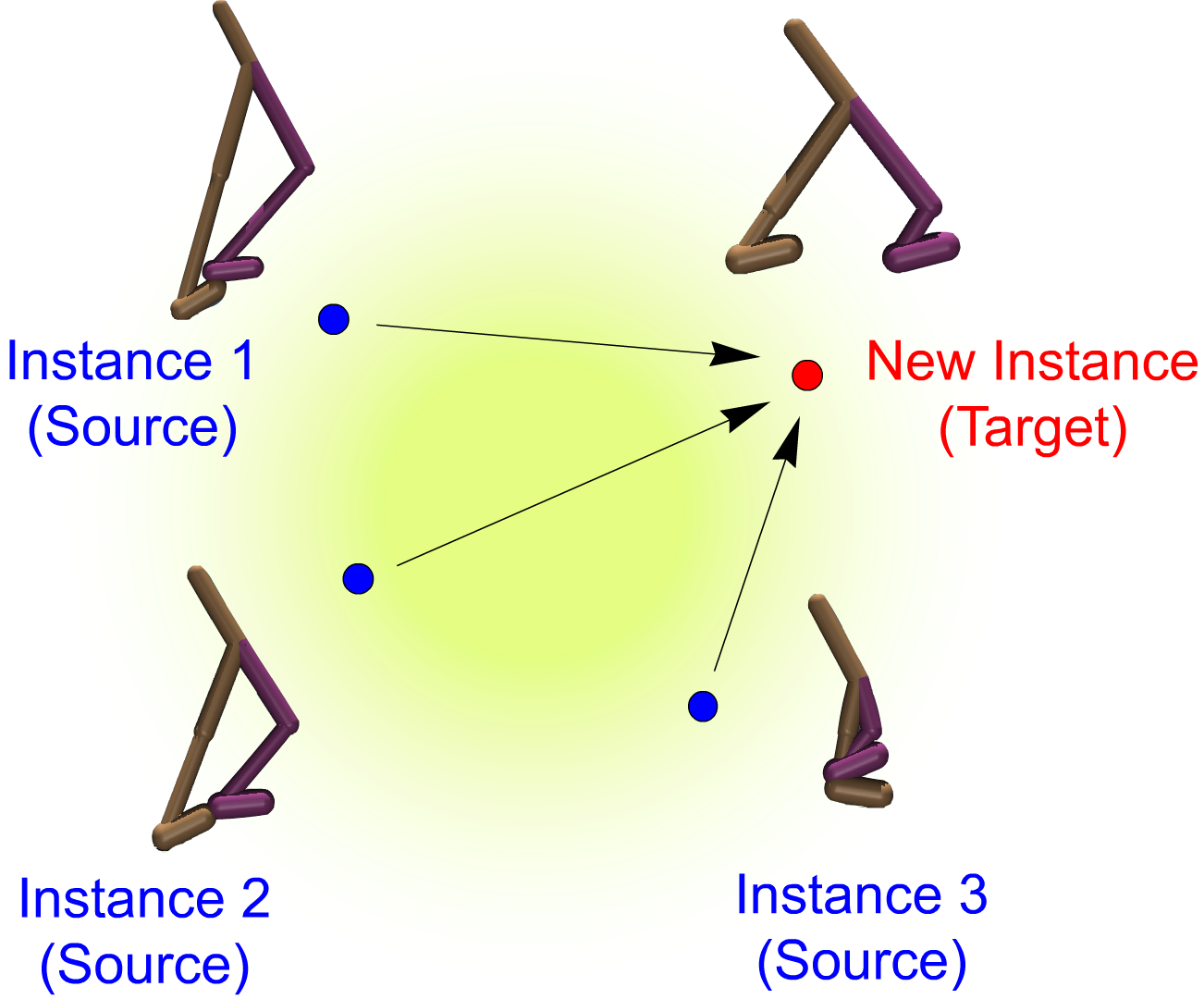}
\caption{Illustration of the CDIL problem studied in the paper. Each dot indicates a robot instance. Our goal is to train a new robot instance~(target) with the existing expert~(source) demonstrations of similar robots. } 
\label{fig:intro_example}
\end{figure}

The availability of demonstration from multiple experts inspires us to mine the invariant representation for behavior patterns, which can be used for CDIL. Taking Walker as an example, one can notice that all these human-like walker instances walk by alternatively moving one leg forward, despite their difference in leg length. Such moving pattern is invariant and shared across the instances. This observation is formalized as the \textit{domain-invariant subspace assumption} in recent work~\cite{transfer_survey}:

\textbf{Assumption} The state space and the action space of similar domains can be disentangled into several independent subspaces~(factors). Some of these subspaces are domain-invariant and shared by source and target tasks. Hence, we can share the task knowledge on these domain-invariant subspaces.

\begin{figure}[t]
\centering
\includegraphics[width=0.95\linewidth]{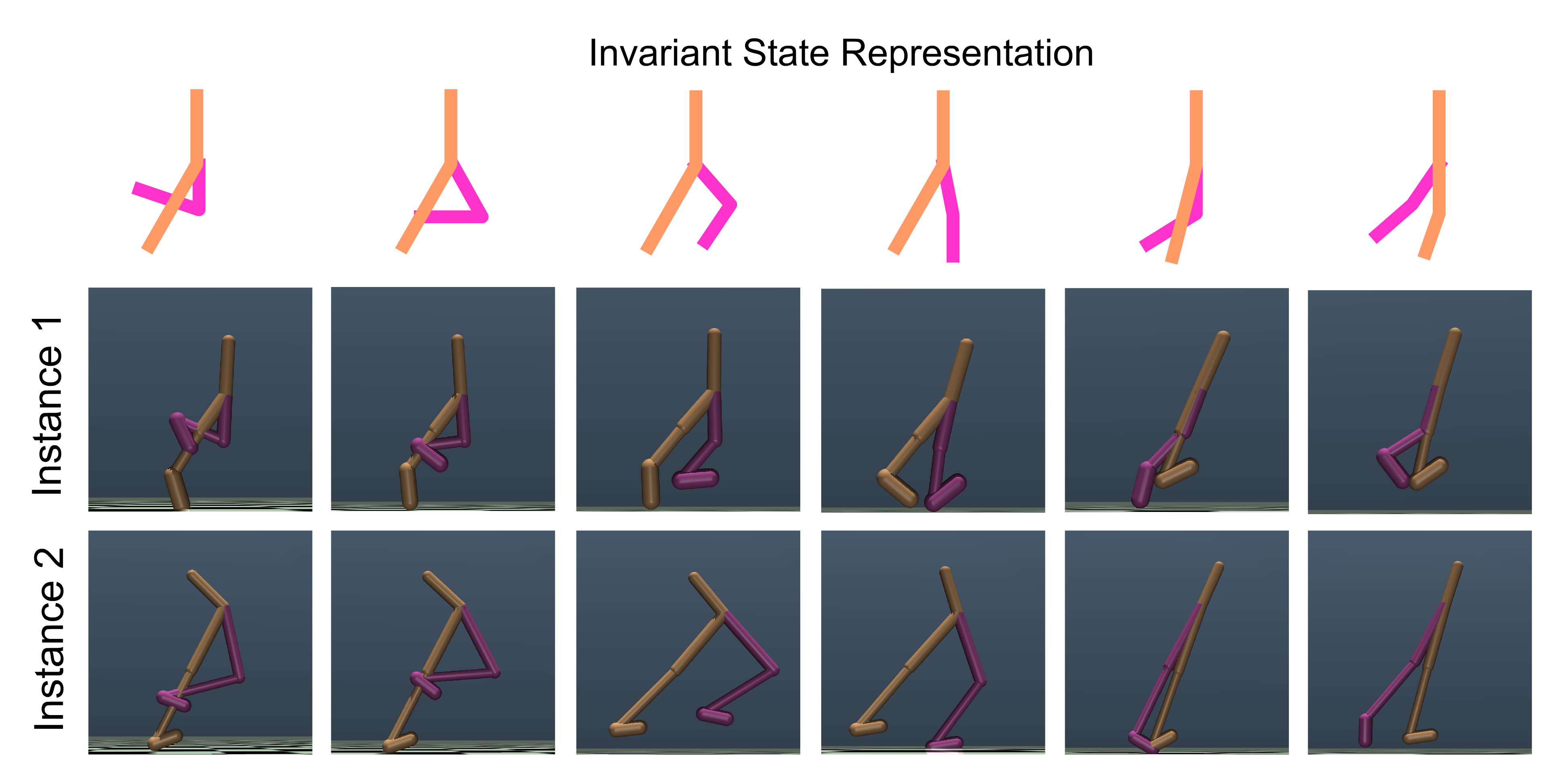}
\caption{Illustration of a possible domain-invariant state representation for Walkers. Despite their physical difference, the behavior of these Walkers can be extracted and described by invariant representation~(above row). We propose a method to learn such invariant representation, and use it to solve the CDIL problem in this paper.} 
\label{fig:intro_inv}
\end{figure}

Our solution directly follows the assumption. We first learn an domain-invariant state and action representation space for robots, and then perform imitation learning inside this invariant representation space. For example, one possible domain-invariant state representation for the Walkers is shown in the Figure~\ref{fig:intro_inv}. We expect that such invariant state representation only encodes their walking behavior, and ignore the irrelevant factors. In other words, this invariant representation can be considered as a behavior prototype for these robots. 

There exist some representation learning methods for similar purpose~\cite{invariant, context_translation}. However, these methods require human-specified paired states from source and target domain. This can be quite inconvenient when the number of experts increases. Moreover, specifying such pairwise states can also be difficult since it is hard to define which states can be paired in some cases~\cite{cross_domain, adapt_dail}. Our proposed method overcomes such problem by leveraging the cycle consistency structure and using domain confusion to align the representation and make it robust.

Our contributions can be summarized as follow:
\begin{itemize}
    \item We study the CDIL problem on a class of similar robots, and propose a new CDIL algorithm based on invariant representation. The proposed algorithm is more convenient since it does not depend on human-labeled pairwise data between source and target task. 
    \item We test the proposed algorithm on various robots in the MuJoCo simulator. Experimental results show that our algorithm can outperform previous methods and generalized to unseen new robots. We also visualize to show that the proposed method is able to learn representations of similar behavior patterns of different robots.
\end{itemize}
\section{Related Work}
\subsection{Imitation Learning}
In order to enable the robots to acquire desired behavior, one basic approach is by Reinforcement Learning~(RL)~\cite{rl}. However, one drawback of applying RL is the difficulty of reward design. To solve this problem, researchers propose behavior acquisition by Imitation Learning~(IL)~\cite{il_survey}. One straightforward approach of IL is Behavioral Cloning~\cite{Behavior_cloning}. Behavioral Cloning solves the IL problem via a supervised process, in which it learn to predict expert-like action given the state. However, Behavioral Cloning algorithms may suffer from the covariate shift problem~\cite{dagger1}. Another line of work is Inverse Reinforcement Learning~(IRL)~\cite{IRL}. IRL algorithms propose to infer the reward function from the expert demonstration. Among IRL algorithms, one recent branch is the Adversarial Imitation Learning~(AIL)~\cite{GAIL, GAIFO}, which trains the agent to match epert's behavior via an adversarial process. 
Compared with Behavioral Cloning, AIL can succeed in various challenging control tasks~\cite{GAIL}. In this work, our imitation learning process follows the AIL framework.

\subsection{Cross Domain Imitation Learning}
One crucial assumption of IL algorithms in the previous part is that the expert~(source) and the agent~(target) should be in the same domain. However, the expert demonstration may come from other similar but different domain in practice. Utilizing numerous existing demonstrations to solve the learning problem on an unseen new domain is quite appealing, and is called CDIL or Adaptive IL~(ADIL)~\cite{adapt_dail, context_translation}. One common ritual of CDIL is to map the states and actions in the source domain to functionally similar states in the agent's domain, which is called domain adaptation~(translation)~\cite{cross_domain}. Such mechanism is also observed in biology~\cite{bio}. Some recent research in robotics directly defines cross domain mapping by human prior knowledge~\cite{xbp} to train a quadruped robot, but the approach is application-specific and can be confined to particular scenarios. Besides, researchers also consider learning such mapping. Some methods learn the mapping by using pairwise data in the source and target domain~\cite{context_translation}. Some other method utilizes auxiliary tasks and dynamics information to align the states and actions without such pairwise data~\cite{adapt_dail}. However, collecting pairwise data or defining tasks on multiple domains can be tricky and difficult. Therefore, these methods can be quite inconvenient in practice. Some vision-based works also propose unsupervised domain translation~\cite{avid}, but such method can misalign states in some cases~\cite{cross_domain}.

We address this problem by learning invariant representations, rather than direct domain translation. Such idea is partially inspired by research in robot transfer learning~\cite{transfer_survey}, whose purpose is quite similar to CDIL. Some previous methods like~\cite{invariant} propose to learn an invariant feature space for transfer learning. However, it requires supervised pairwise data for alignment. Some other methods tackle the CDIL problem by domain confusion~\cite{tpil}, which can also be considered as instances of invariant representation learning. We will discuss these methods in the next section. Learning invariant representations is also related to domain randomization~\cite{domain_randomization}, which uses various domains to make representation used by policy invariant and robust.

One limitation of our method is that it does not consider cross morphology cases~(such as different number of joints). But we believe that this problem can be solved by padding state and action space in the future. Cross morphology CDIL is recently studies in some recent research~\cite{morphology}. 

\subsection{Domain Confusion}  
Some CDIL methods reuse the demonstrations from different robots by getting rid of the domain specific information in the demonstration via representation learning, and this process is usually called domain confusion~\cite{dom_c, dom_c2, domain_im}. The process is implemented by minizing the mutual information between representation and domain label~\cite{tpil, domain_im}. To estimate the mutual information, a mutual information estimator based on neural network called MINE is often used~\cite{MINE}. Besides MINE based implementation, ~\cite{VDB} uses a variational information bottleneck to regularize the mutual information, which is also used in recent domain adaptation method~\cite{VDB_Apply}. This variational information bottleneck can provide an upper bound estimation of mutual information. Some other methods like~\cite{dom_c2, cross_domain, rl_video} also propose to train a domain discriminator for domain confusion. However, we find that merely using domain confusion for CDIL is not optimal in some cases, since it does not use the dynamics information to align the representation. Our method further takes this into consideration.

\section{Background and Preliminaries}
\subsection{Notations}
Since similar robots usually differ from each other only on some particular physical configuration~(parameters) like link length and body mass, in this paper we assume that we can use these physical configuration $c$ to describe a robot, which is denoted as $R_c$. Then, a class of similar robots can be denoted as $RC = \{R_c|c\in\mathcal{C}\}$, where $\mathcal{C}$ is the configuration space (i.e. set of possible physical configuration parameters). For example, we can use $\mathcal{C}_{pend} = [1.5, 2]$ to characterize a class of pendulums whose lengths are between $1.5$m and $2$m. $R_{1.6}$ can represent a pendulum whose length is $1.6$m. A similar environment characterization is also used by~\cite{adail}.

We formalize the control problem of a robot $R_c$ by the Markov Decision Process~(MDP) in RL, which is defined as a tuple $(\mathcal{S}, \mathcal{A}, \mathcal{P}, \mathcal{R})$. $\mathcal{S}$ denotes its state space. $\mathcal{A}$ denotes its action space. $\mathcal{P}$ denotes the transition dynamics from time step $t$ to $t+1$. $\mathcal{R}:\mathcal{S}\times\mathcal{A}\to\mathbb{R}$ is the reward function. At time step $t$, the robot agent observes $s_t\in\mathcal{S}$ and output an action $a_t\in\mathcal{A}$. Then, the environment evolves to $s_{t+1}$ following the dynamics defined by $\mathcal{P}$. Moreover, for the transition~$(s_t, a_t, s_{t+1})$, we also use $c_t$ to denote the corresponding robot's configuration. In this paper, we assume that the dimension of the robots' state space and action space are the same~(the controllable joints are the same across the robots). But note that one state~(action) can be of different meanings to different robots. For example, let the state space of a pendulum is the tilting angle~(for simplicity). Then pendulums of different length should react differently to the same observed tilting angle. This is due to the difference of dynamics of different robots. 

\subsection{Generative Adversarial Imitation Learning}
We use Generative Adversarial Imitation Learning~(GAIL)~\cite{GAIL} as our imitation learning framework. GAIL enables the agent to obtain expert-like behavior by encouraging it to match its behavior with expert by fooling a discriminator network $D$. Such discriminator $D$ is trained to discern expert $\pi_E$'s state-action pair from agent $\pi$'s state-action pair, which is equivalent to minimizing the following loss:
$$
    - \mathbb{E}_{\pi_E}[\log (D(s_t, a_t))] - \mathbb{E}_{\pi}[\log (1 - D(s_t, a_t))].
$$
Then, we can use such discriminator $D$ to define the imitation reward for the agent. The reward for taking $a_t$ at state $s_t$ is defined as $r_t = -\log(1 - D(s_t, a_t))$.

\subsection{Mutual Information Neural Estimation}
Our algorithm requires mutual information estimation to learn robust invariant representation. To evaluate the mutual information, we apply the estimation method proposed by MINE~\cite{MINE}. This method follows from the Donsker Varadhan representation theorem, which states that given two random variables $X$ and $Z$ on the sample space $\Omega$, the mutual information between $X$ and $Z$ is given by
$$
    I(X, Z) = \sup\limits_{T:\Omega\to\mathbb{R}} \mathbb{E}_{x\sim P_{XZ}} T(x) - \log\mathbb{E}_{x\sim P_XP_Z }e^{T(x)}.
$$
In the above equation, $P_{XZ}$ denotes the joint probability distribution of $X$ and $Z$, $P_XP_Z$ denotes the product of their marginal distributions. The supremum is taken over functions such that the expectation terms are finite. MINE proposes to parameterize the mapping $T$ with a neural network $T_\phi$, so
$$
    I_T(X, Z) = \mathbb{E}_{x\sim P_{XZ}} T(x) - \log\mathbb{E}_{x\sim P_XP_Z }e^{T(x)}
$$
provides a lower bound estimation of $I(X, Z)$. We can get an approximation of $I(X, Z)$ by maximizing $I_{T_\phi}$ and calculate its value. 
\begin{figure}[ht]
\centering
\includegraphics[width=0.9\linewidth]{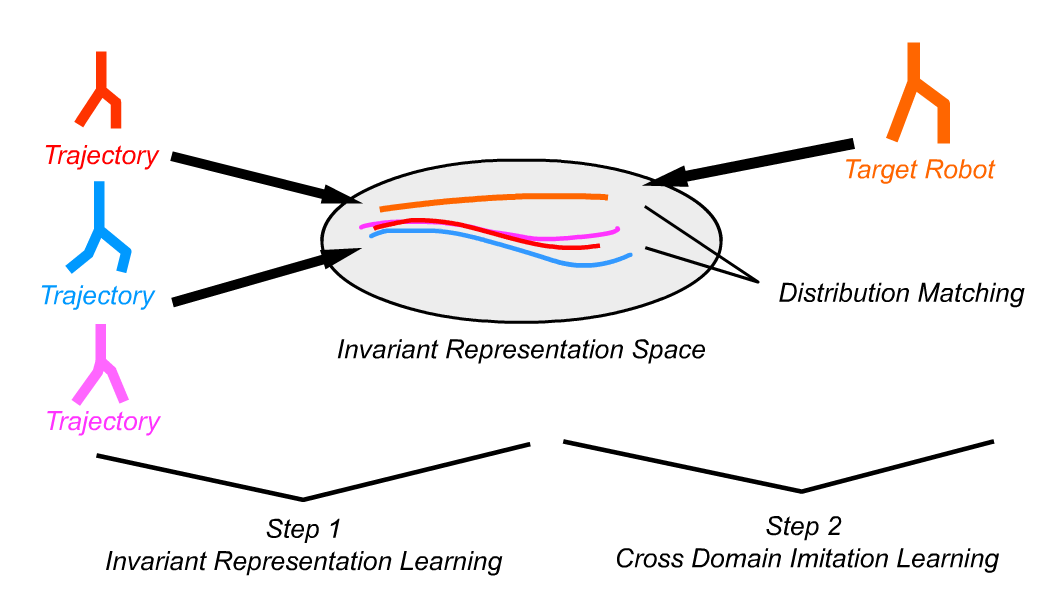}
\caption{Overall process of the proposed algorithm. The first step is to train an invariant representation module. The second step is cross domain imitation learning on the target robot. We train the target robot to match expert's behavior in the invariant representation space.}
\label{fig:methods}
\end{figure}
\section{Algorithm}
\subsection{Overview}
\begin{figure*}[ht]
\centering
\includegraphics[width=0.9\linewidth]{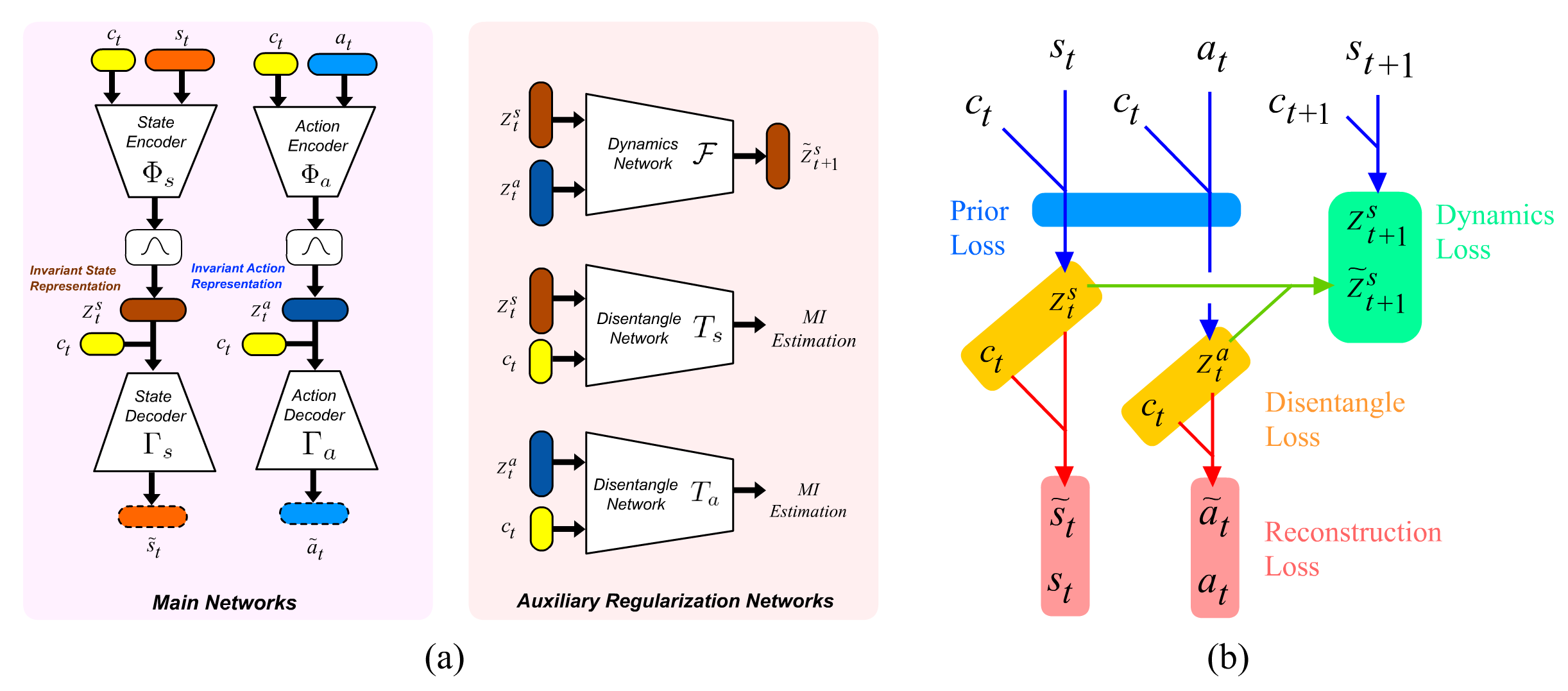}
\caption{(a) Structure of the proposed invariant representation module. (b) Illustration of the loss functions.}
\label{fig:network}
\end{figure*}
We first provide the overview of our proposed learning process in Figure~\ref{fig:methods}. The whole process is composed of two steps. The first step is to train a invariant representation module. This module maps the states and actions of each robot instance to the an invariant representation space where similar states and actions in different robots are aligned. In the second step, we carry out imitation learning. We transform the state-action pairs of the target robot into the invariant representation space, and use GAIL to train the target robot to match them with the expert demonstration. 
\subsection{Invariant Representation Module}
We illustrate the invariant representation module in the Figure~\ref{fig:network}. It has two parts: the main part consists of encoding-decoding networks, which produce invariant representation. The other part consists of some auxiliary regularization networks to further improve the representation. 
\subsubsection{Main Network}
We use two variational encoders to encode states and actions respectively. Concretely, we define the state encoder as $\Phi_s(\cdot|s_t, c_t)$, which takes current state $s_t$ and robot configuration $c_t$ as input, and produces a Gaussian distribution on the invariant state space $\mathcal{Z}_S$. The encoded invariant state representation $z^c_t$ is then sampled from $\Phi_s(\cdot|s_t, c_t)$. The action encoder $\Phi_a(\cdot|a_t, c_t)$ is defined on the invariant action space $\mathcal{Z}_A$ in the same way. We sample invariant action representation $z^a_t\sim\Phi_a(\cdot|a_t, c_t)$. Then, we use two decoders to ensure that the representation contains enough information about the input. The state decoder $\Gamma_s(z_t^s, c_t)$ is required to reconstruct robot's state $s_t$ conditioned on $z_t^s$ with its configuration $c_t$. The action decoder $\Gamma_a(z_t^a, c_t)$ is defined similarly. The reconstructed state and action are denoted as $\tilde{s}_t$ and $\tilde{a}_t$. Then the reconstruction loss functions for state and action representation are defined as
\begin{equation}
    \mathcal{L}_{sr} = \mathbb{E}_{\Phi_s(z|s_t, c_t)}[\Vert s_t - \tilde{s}_t\Vert^2]
\end{equation}
and
\begin{equation}
   \mathcal{L}_{ar} = \mathbb{E}_{\Phi_a(z|a_t, c_t)}[\Vert a_t - \tilde{a}_t\Vert^2]
\end{equation}
respectively. To enforce Gaussian distribution, a prior loss is also used. It is defined as
\begin{equation}
    \mathcal{L}_{kl} = \mathcal{L}_{skl} + \mathcal{L}_{akl},
\end{equation}
where
\begin{align}
    \mathcal{L}_{skl} &= D_{KL}(\Phi_s(z|s_t, c_t)||p(z)), \\
    \mathcal{L}_{akl} &= D_{KL}(\Phi_s(z|s_t, c_t)||p(z)).
\end{align}
Here, $D_{KL}$ is the Kullback-Leibler divergence. $p(z)$ is the prior Gaussian distribution $\mathcal{N}(0, 1)$.

\subsubsection{Dynamics Loss}
As is introduced before, similar robot instances share their motion~(dynamics) pattern in the invariant representation space. Therefore, we also train a dynamics model in the representation space for all the robot instances to discover shared pattern. Such dynamics model is a neural network $\mathcal{F}:\mathcal{Z}_S\times\mathcal{Z}_A\to\mathcal{Z}_S$, whose goal is to predict the representation of the future state given the current state and action. This prediction is denoted as $\tilde{z}_{t+1}^s = \mathcal{F}(z_t^s, z_t^a)$. The loss function is defined as
\begin{equation}
    \mathcal{L}_{dyn} = \mathbb{E}\Vert \tilde{z}_{t+1}^s - z_{t+1}^s \Vert^2 = \mathbb{E}\Vert \mathcal{F}(z_t^s, z_t^a) - z_{t+1}^s)\Vert^2.
\end{equation}
Such loss function is actually a cycle consistency constraint between the original state space and representation space. \cite{cross_domain} finds that such constraint can push similar states (actions) towards each other, which is also essential for our problem.

\begin{table*}[bp]
\centering
\caption{The Interpolation~(Int.) and Extrapolation~(Ext.) Performance of Evaluated Algorithms.}
\renewcommand{\arraystretch}{1.3}
\begin{tabular}{|l|l|p{1.5cm}|p{1.3cm}|p{1.3cm}|p{1.3cm}|p{1.3cm}|p{1.3cm}|p{1.55cm}|p{1.35cm}|}
\hline
Mode                         & Algorithm & IPendulum~(K) & IDPendulum & Swimmer   & Hopper    & Walker    & Cheetah~(K)   &IPendulum~(A)& Cheetah~(A)\\ \hline
\multirow{3}{*}{Int.} & GAIL      & 1.00$\pm$0.00 & 0.82$\pm$0.26  & 0.72$\pm$0.10 & 0.90$\pm$0.02 & 0.58$\pm$0.04 & 0.49$\pm$0.25 &1.00$\pm$0.00&0.76$\pm$0.09\\ \cline{2-10} 
                             & TPIL      & 1.00$\pm$0.00 & 0.85$\pm$0.13 & 0.86$\pm$0.05  & 0.93$\pm$0.02          & 0.64$\pm$0.04          & 0.58$\pm$0.21           &1.00$\pm$0.00&0.85$\pm$0.04\\ \cline{2-10} 
                             & IR-GAIL   & \textbf{1.00$\pm$0.00}  & \textbf{0.96$\pm$0.04} & \textbf{0.98$\pm$0.02} & \textbf{0.99$\pm$0.01}  & \textbf{0.83$\pm$0.08} &   \textbf{0.91$\pm$0.04} & \textbf{1.00$\pm$0.00} &\textbf{0.86$\pm$0.06}\\ \hline
\multirow{3}{*}{Ext.} & GAIL      & 0.53$\pm$0.52   & 0.04$\pm$0.01      & 0.65$\pm$0.11 & 0.78$\pm$0.05  & 0.45$\pm$0.12   & 0.22$\pm$0.14 &1.00$\pm$0.00&0.71$\pm$0.10\\ \cline{2-10} 
                             & TPIL      &  0.78$\pm$0.43 & 0.05$\pm$0.01  & 0.77$\pm$0.06  & 0.80$\pm$0.04 & 0.57$\pm$0.06 & 0.46$\pm$0.15  &1.00$\pm$0.00&\textbf{0.81$\pm$0.08}\\ \cline{2-10} 
                             & IR-GAIL   &  \textbf{1.00$\pm$0.00} & \textbf{0.72$\pm$0.18}  & \textbf{0.91$\pm$0.03}  & \textbf{0.92$\pm$0.02} & \textbf{0.71$\pm$0.06} & \textbf{0.80$\pm$0.21}  & \textbf{1.00$\pm$0.00} &0.79$\pm$0.11\\ \hline
\end{tabular}
\renewcommand{\arraystretch}{1.0}
\end{table*}

\subsubsection{Disentangle Loss}
A necessary condition to make representation invariant is that the representation of different robots should be indistinguishable, otherwise the representation must carry robot-specific information. This condition is equivalent to minimizing the mutual information between encoded representation and robot's configuration. In our setting, both $I(z_t^s, c_t)$ and $I(z_t^a, c_t)$ should be minimized. Hence we train two MINE networks $T_s$ and $T_a$ to provide estimation for $I(z_t^s, c_t)$ and $I(z_t^a, c_t)$, which are written as $I_{T_s}(z_t^s, c_t)$ and $I_{T_a}(z_t^a, c_t)$ respectively. Then we define the disentangle loss for encoders as
\begin{equation}
    \mathcal{L}_{disent} = I_{T_s}(z_t^s, c_t) + I_{T_a}(z_t^a, c_t).
\end{equation}
The loss function for updating $T_s$ and $T_a$ is defined as
\begin{equation}
    \mathcal{L}_{T} = -(I_{T_s}(z_t^s, c_t) + I_{T_a}(z_t^a, c_t)).
\end{equation}
Our final training loss is defined as a weighted combination of the above loss functions:
\begin{equation}
    \mathcal{L} = \mathcal{L}_{sr} + \mathcal{L}_{ar} + \lambda_1\mathcal{L}_{disent} + \lambda_2\mathcal{L}_{dyn} + \lambda_3\mathcal{L}_{kl}.
\end{equation}
Here, $\lambda_1$, $\lambda_2$ and $\lambda_3$ are weighting hyperparameters. To obtain the training data, we sample transition data $(s_t, a_t, s_{t+1})$ on various robot instances in the given robot family using random policy. However, applying random policy for sampling may leave some important transitions uncovered in some tasks. Therefore, we also add rollout transition data of experts into training dataset if such situation happens. This is enough as expert rollouts can cover all the important transitions. Since the encoders $\Phi_s$ and $\Phi_a$ are evolving during the training process, we update $T_a$ and $T_s$ accordingly to track the change of the mutual information. In the implementation, we update network $T_s$ and $T_a$ by minimizing $\mathcal{L}_T$ after each update step of encoders. 

\subsection{Imitation Learning with Invariant Representation}

After learning the invariant representation, we use it for imitation in the GAIL fashion. The only difference from GAIL is that we use the encoded invariant states and actions in the calculation. So the loss function for the discriminator is defined as
\begin{equation}
    \mathcal{L}_{d} = - \mathbb{E}_{\pi_E}[\log (D(z_t^s, z_t^a)] - \mathbb{E}_{\pi}[\log (1 - D(z_t^s, z_t^a)].
\end{equation}
The reward for the agent $R_c$ is defined as $r_t = -\log (1 - D(z_t^s, z_t^a))$. The overall imitation learning process is summarized in Algorithm~\ref{alg:overall}.
\begin{algorithm}
    \caption{IR-GAIL}
    \label{alg:overall}
        \KwIn{Policy $\pi$ and discriminator $D$. Agent configuration $c$}
        \tcp{Step 1: Representation Learning}
        Build training dataset by random rollout and expert demo.\;
        \For{$i=1,2,...$}{
            Optimize $\Phi_s, \Phi_a, \Gamma_s, \Gamma_a, \mathcal{F}$ by minimizing $\mathcal{L}$\;
            Optimize $T_s$, $T_a$ by minimizing $\mathcal{L}_T$\;
        }
        \tcp{Step 2: Imitation Learning}
        \For{$i=1,2,...$}{
            Sample trajectories using $\pi_{\theta_i}$\;
            Update $D$ by minimizing $\mathcal{L}_{d}$\;
            Update $\pi$ based on $r_t$ using PPO~\cite{PPO}\;
        }
\end{algorithm}

\section{Experiments}
\subsection{Settings}
We validate our algorithm on several MuJoCo control benchmarks. We use InvertedPendulum, InvertedDoublePendulum, Hopper, Walker, Swimmer, and Halfcheetah. We setup the configuration space $\mathcal{C}$ for these robots as follow. The values shown below are relative to the default configuration value in MuJoCo.
\subsubsection{InvertedPendulum} The configuration space is $[0.75, 5.0]\times[0.5, 2.0]$. The first dimension corresponds to the length of the link. The second dimension corresponds to the maximum gear. 
\subsubsection{InvertedDoublePendulum} The configuration space is $[1.0, 3.0]\times[1.0, 3.0]\times[0.5, 1.5]$. The first and the second dimension are the length of the bottom link and the above link respectively. The third dimension is the weight of the above link. 
\subsubsection{Hopper} The configuration space is $[0.5, 3.0]\times[0.5, 3.0]$. The first and the second dimension correspond to the length of the body and the length of the thigh respectively.
\subsubsection{Walker} The configuration space is $[0.5, 2.0]\times[0.5, 2.0]$. The first and the second dimension correspond to the length of the thigh and the shank respectively.
\subsubsection{Swimmer} The configuration space is $[0.5, 2.0]\times[0.5, 2.0]\times[0.5, 2.0]$. The first, second, and third dimension correspond to the length of the head, body, and the tail respectively.
\subsubsection{Cheetah} The configuration space is $[0.5, 2.5]\times[0.5, 2.0]\times[0.5, 2.0]$. The first and the second dimension correspond to the length of the body and back leg respectively. The third dimension correspond to the maximum gear of the back leg. 

We use the default sensor data as state observation in the experiments. For the InvertedPendulum and Cheetah, we use two different types of state observation. The first type is based on the body keypoints~(K). The second type is based on the joint angles~(A), which is the same as the default MuJoCo setup.  

\subsection{Evaluation}
We evaluate the performance of algorithms by interpolation and extrapolation experiments. In the interpolation experiment, we sample robots whose physical configurations are quite close to the experts'. In the extrapolation experiment, however, the configurations of sampled robots differ significantly from the experts'. 
For each environment, the expert demonstrations and random rollout used for training are collected on robots sampled from a ball region $\mathcal{B}$ in the configuration space, so that we can sample robots in~(out of) $\mathcal{B}$ for interpolation~(extrapolation) evaluation. We collect the random rollout on 16 random sampled robots. We sample 32 trajectories from 4 different experts for imitation. For policy optimization, we use PPO with Generalized Advantage Estimation~\cite{gae}. 

\begin{figure}[t]
\centering
\includegraphics[width=0.9\linewidth]{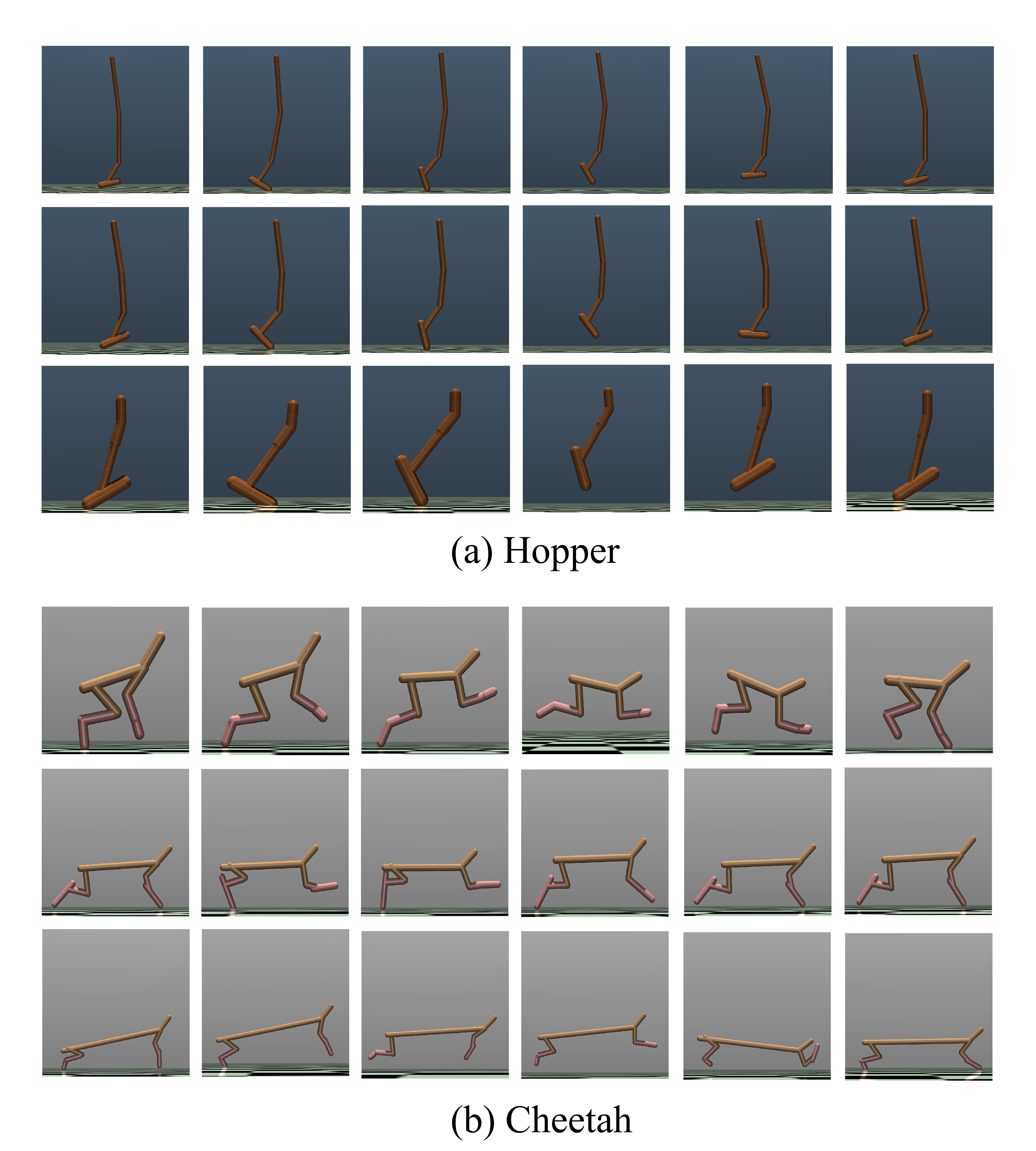}
\caption{Qualitative coupling results. Each row corresponds to states of one robot instance. Each column shows a group of coupled states. The coupled states in the same group are functionally similar.}
\label{fig:visualization}
\end{figure}
\subsection{Results}
We compare the proposed IR-GAIL with GAIL, and another domain confusion based CDIL algorithm called TPIL~\cite{tpil}. The input of TPIL is fully observational, so we have to adapt it to our setting. The result is shown in the Table 1. We calculate the mean and standard deviation of each algorithm's return, which are normalized  with respect to the average return of the expert policy and the random policy. We can see that our proposed IR-GAIL in general performs better than GAIL and TPIL, though these methods can still learn from the expert on some of these domains, which is due to the similarity of observations and dynamics. IR-GAIL can also generalize well to robots out of the training distribution, though the extrapolation performance of IR-GAIL is lower than interpolation performance. We also notice that the previous algorithms can still perform quite well on InvertedPendulum and Cheetah, if the observation is based on joint angle. Such joint angle based observation is naturally invariant representations in some cases. For example, if the robots are only differerent from each other in size, then the angle information is still useful among them. However, measuring the angle accurately in application can be difficult, and keypoint based observation is used more frequently. 

\subsection{Ablation Studies}
We also test the performance of the model after removing the dynamics regularization.  We find that removing such regularization will not lead to a performance drop on InvertedPendulum and Hopper. However, we observe the performance on Walker, Swimmer, Cheetah, and InvertedDoublePendulum drops by 12\%, 19\%, 26\%, and 43\% respectively in average after its removal. An interesting fact is that the performance after such removal is still higher than the baselines in general. The reason is that variation bottleneck and the disentangle loss naturally regularize the representation and make it robust, which is also reported by~\cite{domain_im}.

\subsection{Qualitative Results}
In this part, we provide some qualitative coupling results to understand the learned invariant representation space in Figure~\ref{fig:visualization}. We first rollout trajectories of different robot instances by trained policies, and map the encountered states into the invariant representation space using the trained invariant representation module. To obtain a group of coupled states of different robot instances, we sample a point in the representation space and find its nearest representations of different robot instances. Then the corresponding robot states of these representations form a group of coupled states. The results are collected from Hopper and Cheetah. We can find that the coupled states in each group are highly functionally similar. This result suggests that our method can encode desired invariant representation for similar robots. 

\section{Conclusion}
In this paper, we studied the CDIL problem on a class of similar robots, and propose a new CDIL algorithm based on invariant representation. Experimental results showed that our method can achieve superior performance. We used visualization to demonstrate that our method can learn invariant representation for CDIL. One limitation of this work is that it is not observational, and we will study how to extend this approach to observational IL in the future. Another future direction is to study how to infer robot's configuration automatically. We will also try to apply the proposed method on real robots.

\balance
\bibliographystyle{abbrv}
\bibliography{iscv}
\end{document}